\documentclass[conference]{IEEEtran}
\IEEEoverridecommandlockouts
\usepackage{cite}
\usepackage{amsmath,amssymb,amsfonts}
\usepackage{algorithmic}
\usepackage{graphicx}
\usepackage{textcomp}
\usepackage{xcolor}
\def\BibTeX{{\rm B\kern-.05em{\sc i\kern-.025em b}\kern-.08em
    T\kern-.1667em\lower.7ex\hbox{E}\kern-.125emX}}
\usepackage{bm}
\usepackage{url}
\usepackage{fancyvrb} % Provides the enhanced Verbatim environment
\usepackage{threeparttable}
\usepackage{tabularx}
\usepackage{hyperref}
\usepackage{verbatim} % For verbatim environments
\usepackage{tikz}
\usepackage{fancyvrb}
\usepackage{booktabs}
\hypersetup{
  colorlinks = true, % enable colored links
  linkcolor = blue,  % color of internal links
  citecolor = blue, % color of citation links
  filecolor = magenta, % color of file links
  urlcolor = cyan    % color of external links
}
\usetikzlibrary{arrows.meta, positioning}
\def\BibTeX{{\rm B\kern-.05em{\sc i\kern-.025em b}\kern-.08em
    T\kern-.1667em\lower.7ex\hbox{E}\kern-.125emX}}

\usepackage{tikz}
\usetikzlibrary{shapes.geometric, arrows, positioning}

\tikzstyle{startstop} = [rectangle, rounded corners, minimum width=3cm, minimum height=1cm,text centered, draw=black, fill=red!30]
\tikzstyle{process} = [rectangle, minimum width=3cm, minimum height=1cm, text centered, draw=black, fill=blue!30]
\tikzstyle{decision} = [diamond, minimum width=3cm, minimum height=1cm, text centered, draw=black, fill=green!30]
\tikzstyle{arrow} = [thick,->,>=stealth]
    
\begin{document}

\title{Mapping Biomedical Ontology Terms to IDs: Effect of Domain Prevalence on Prediction Accuracy\\
}

\author{\IEEEauthorblockN{Thanh Son Do}
\IEEEauthorblockA{\textit{Computer Science Department} \\
\textit{Missouri State University}\\
Springfield, MO \\
td64s@MissouriState.edu\\}
\and
\IEEEauthorblockN{Daniel B. Hier}
\IEEEauthorblockA{\textit{Kummer Institute Center for AI and AS %Autonomous Systems
} \\
% \IEEEauthorblockA{\textit{and Autonomous Systems} \\
\textit{Missouri University of Science \& Technology}\\
Rolla, MO \\
hierd@umsystem.edu}
\and
\IEEEauthorblockN{Tayo Obafemi-Ajayi}
\IEEEauthorblockA{\textit{Cooperative Engineering Program} \\
\textit{Missouri State University}\\
Springfield, MO \\
tayoobafemiajayi@missouristate.edu}
% \and
% \and
% \and
}

\maketitle
\begin{abstract}
This study evaluates the ability of large language models (LLMs) to map biomedical ontology terms to their corresponding ontology IDs across the Human Phenotype Ontology (HPO), Gene Ontology (GO), and UniProtKB terminologies. Using counts of ontology IDs in the PubMed Central (PMC) dataset as a surrogate for their prevalence in the biomedical literature, we examined the relationship between ontology ID prevalence and mapping accuracy. Results indicate that ontology ID prevalence strongly predicts accurate mapping of HPO terms to HPO IDs, GO terms to GO IDs, and protein names to UniProtKB accession numbers. Higher prevalence of ontology IDs in the biomedical literature correlated with higher mapping accuracy. Predictive models based on receiver operating characteristic (ROC) curves confirmed this relationship.

In contrast, this pattern did not apply to mapping protein names to Human Genome Organisation's (HUGO) gene symbols. GPT-4 achieved a high baseline performance (95\%) in mapping protein names to HUGO gene symbols, with mapping accuracy unaffected by prevalence. We propose that the high prevalence of HUGO gene symbols in the literature has caused these symbols to become \textit{lexicalized}, enabling GPT-4 to map protein names to HUGO gene symbols with high accuracy.
These findings highlight the limitations of LLMs in mapping ontology terms to low-prevalence ontology IDs and underscore the importance of incorporating ontology ID prevalence into the training and evaluation of LLMs for biomedical applications.
\end{abstract}

\section{Introduction}
\IEEEPARstart{W}ith the growing availability of comprehensive biomedical ontologies \cite{bodenreider2005biomedical,rubin2008biomedical,konopka2015biomedical}, the task of ontology mapping has gained importance in biomedical informatics. Ontology mapping is the linking of unstructured textual data to structured ontology terms and their ontology IDs. These mappings enable data exchange and interoperability and the downstream use of data in research, precision medicine \cite{robinson2015capturing}, clinical decision-making \cite{sox2024medical}, and population health \cite{young2004population}.

\begin{figure}[t!]
    \centering
    \begin{tikzpicture}[
        node distance=1.2cm and 2cm, % Distance between nodes
        every node/.style={draw, text width=3cm, align=center, font=\small, rounded corners}, % Rounded box style
        arrow/.style={-{Latex}, thick} % Arrow style
    ]
        % Nodes
        \node (source) {Clinical Note\\(Unstructured Text)};
        \node (extract) [below=of source] {\textbf{Term Extraction}\\(e.g., ''impaired reflexes")};
        \node (standardize) [below=of extract] {\textbf{Term Standardization}\\(e.g., ''hyporeflexia")};
        \node (map) [below=of standardize] {\textbf{Ontology Mapping}\\(e.g., HP:0001265)};

        % Arrows with process descriptions
        \draw [arrow] (source) -- (extract) node[midway, right] {Term Identification};
        \draw [arrow] (extract) -- (standardize) node[midway, right] {Convert to Standard Term};
        \draw [arrow] (standardize) -- (map) node[midway, right] {Match Term to Ontology ID};
   
    \end{tikzpicture}
    \caption{\textbf{Workflow for Biomedical Term Normalization.}~The process extracts terms from clinical notes, standardizes them to ontology terms, and maps them to ontology identifiers (e.g., HP:0001265).}
    \label{fig:biomedical_workflow}
\end{figure}

Ontology mapping is performed in three sequential steps \cite{yehia2019ontology, dahdul2018annotation, groza2015automatic, krauthammer2004term, funk2014large}: \textit{term identification}, \textit{term standardization}, and \textit{ontology ID linking} (Fig. \ref{fig:biomedical_workflow}). While large language models (LLMs), such as GPT-4, excel at term identification and standardization, they often face challenges in accurately mapping terms to their corresponding ontology IDs \cite{groza2024evaluation}. These challenges stem from the stochastic nature of LLMs, limited exposure to rare ontology IDs during pretraining, and the absence of a built-in lookup mechanism. For instance, despite generating detailed and accurate definitions, GPT-4 may confabulate or hallucinate ontology IDs for rare terms \cite{groza2024evaluation}. These errors indicate that LLM performance may rely on the prevalence of ontology terms and their IDs in both the training data and the biomedical literature. Based on these observations, we hypothesize that ontology IDs with higher frequencies in the biomedical literature are more likely to be accurately mapped to their corresponding terms by LLMs.

In this study, we evaluate the ability of GPT-4 (Generative Pre-Trained Transformer) to perform ontology ID mapping across three distinct terminologies: Human Phenotype Ontology (HPO) terms \cite{kohler2017human} to their corresponding HPO IDs, Gene Ontology (GO) terms \cite{gene2019gene} to their associated GO IDs, and protein names from the UniProtKB (curated human proteins subset) \cite{uniprot2015uniprot} to their accession numbers (AC) and HUGO gene symbols (GS) \cite{eyre2006hugo}.
\section{Methods}
\subsection*{Data Acquisition}
To ensure robustness and generalization of evaluation results, ontology terms and their IDs were retrieved from publicly available resources:
\begin{itemize}
    \item \textbf{HPO}: Terms and IDs were downloaded from the NCBO website \cite{HPO} as a CSV file.
    \item \textbf{GO}: Cellular component (CC) terms and IDs were retrieved from the NCBO website \cite{GO} as a CSV file.
    \item \textbf{UniProtKB}: Protein names, gene symbols, and accession numbers were extracted from \textit{uniprot.sprot.dat} (\url{https://www.uniprot.org/help/downloads}).
\end{itemize}

A total of 18,800 HPO ontology terms, 1,839 GO ontology terms (CC hierarchy), 3,979 protein names (for accession numbers), and 3,974 protein names (for gene symbols) were selected for evaluation.

\subsection*{Mapping Ontology Terms to Ontology IDs}
Ontology terms were inputted to GPT-4 via the OpenAI API. GPT-4 was prompted to retrieve the corresponding ontology IDs. The following four distinct mapping scenarios were evaluated:

\vspace{0.5cm} % Adjust the value as needed to add space

\begin{tabular}{lll}
\textbf{Mapping Input} & &\textbf{Mapping Output}\\
HPO ontology term &$\mapsto$ &HPO ontology ID \\
GO ontology term (CC) &$\mapsto$& GO ontology ID \\
Protein name &$\mapsto$& UniProtKB access. num \\
Protein name &$\mapsto$& HUGO gene symbol \\
\end{tabular}
\vspace{0.2cm} % Adjust the value as needed to add space

Mapping accuracy is defined as the proportion of terms mapped to their correct ontology IDs. To investigate the role of term and ID prevalence counts for terms and IDs were obtained from the PMC full-text database via the PMC API at:
\url{https://eutils.ncbi.nlm.nih.gov/entrez/eutils/esearch.fcgi}. Counts (dataset-wide)  were stored as dataframes for analysis. 

\subsection*{Data Analysis}
% Analyses were performed using Python and scikit-learn. 
To systematically evaluate the relationship between domain prevalence and mapping accuracy across each ontology, we performed the following analyses for each mapping task:
% To systematically investigate the impact of term prevalence on mapping accuracy

\begin{itemize}
    \item Correlation coefficients (Pearson $r$) were calculated between mapping accuracy and PMC frequencies of ontology terms and IDs.
    \item The ontology terms were rank-ordered based on the counts of their ontology IDs and subsequently partitioned into 20 equal-sized bins. Mean mapping accuracy and ID counts were calculated per bin and plotted.
    \item  Log-transformed rank-frequency distributions (Zipf plots) \cite{zipf2016human, li2002zipf} were created for ontology IDs, with jitter applied to prevent marker overlap, ensuring better visualization of ID distribution and mapping accuracy trends. To explore the relationship between ID counts and mapping accuracy, markers on the Zipf plots were color-coded according to their mapping accuracy.
\end{itemize}

\begin{table}[ht!]
\centering
\caption{Performance Metrics for Different Ontology Mappings}
\label{tab:performance_metrics}
\begin{tabular}{lrrrr}
\toprule
Map to $\rightarrow$ & \textbf{HPO ID} & \textbf{GO ID} & \textbf{AC} & \textbf{GS} \\
\midrule
\textbf{Metrics}&&&&\\
{Terms to Map (N)} & 18,800 & 1,839 & 3,979 & 3,974 \\
{Mean ID PMC Count} & 1 & 17 & 29 & 11,366 \\
{Mean Term PMC Count} & 22,255 & 72,887 & 19,016 & 19,026 \\
% {Baseline Accuracy (\%)} & 8 & 35 & 46 & 95 \\
{{Baseline Accuracy (\%)}} & 8 & 67 & 53 & 95\\
{Correct Mappings to ID} & 1,504 & 599 & 1,830 & 3,775 \\
\bottomrule
\end{tabular}
\begin{tablenotes} % Begin the tablenotes environment
\scriptsize
    \item [a] Table summarizes mapping performance for four different mappings: HPO term to HPO ID, GO term from cellular component hierarchy (CC) to GO ID, protein name to UniProtKB accession number (AC), and protein name to HUGO gene symbol (GS). 
\end{tablenotes}
\end{table}

\begin{table}[ht!]
\centering
\caption{ROC and Correlation Metrics for Ontology Term Mapping Accuracy}
\label{tab:ROC and Correlation Metrics}
\begin{tabular}{lrrrr}
\toprule
Term Maps to $\rightarrow$ & \textbf{HPO ID} & \textbf{GO ID} & \textbf{AC} & \textbf{GS} \\
\midrule
\multicolumn{5}{l}{{\textbf{ROC Curve}}} \\
AUC for ROC Curve & 0.83 & 0.90 & 0.78 & 0.64 \\
Threshold by ROC Curve\tnote{a} & 2 & 4 & 12 & 746 \\
\midrule
\textbf{Correlations} &  &  &  &  \\ 
$r$(ID count, Accuracy) & \textbf{0.33} & \textbf{0.31} & \textbf{0.32} & 0.00 \\ 
$r$(Term count, Accuracy) & 0.03 & 0.14 & 0.03 & 0.00 \\
\bottomrule
\end{tabular}
\begin{tablenotes}
\scriptsize
    \item[a] ROC thresholds represent the optimal balance between sensitivity and specificity for ontology term mapping.
    \item[b] Correlations (r) are calculated between ID count in PMC (or term count in PMC) and mapping accuracy. PMC refers to the PubMed Central full-text database.
\end{tablenotes}
\end{table}

\begin{figure}[h!]
    \centering
    \includegraphics[width=0.95\linewidth]{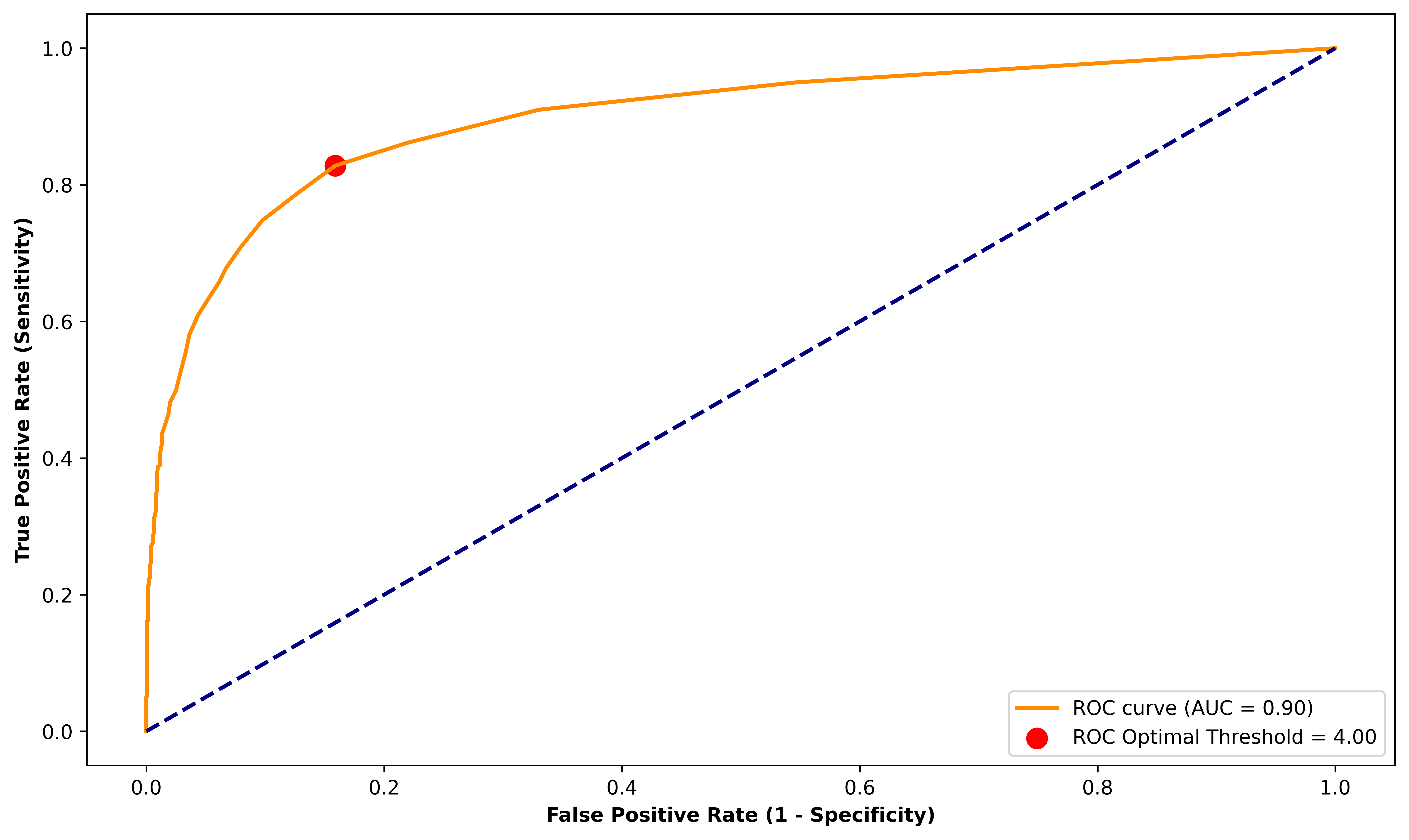}
    \caption{{\textbf{ROC Curve for Mapping GO Terms to GO IDs.} 
    Using 1,839 GO terms from the Cellular Component (CC) hierarchy, the ROC curve (orange line) was computed to evaluate mapping accuracy. An optimal threshold of 4 ID counts in the PMC dataset (red circle) maximized sensitivity and specificity, achieving an AUC of 0.90. Similar curves were created for the HPO and UniProtKB terminologies (not shown).}}
    \label{fig:ROC_GO}
\end{figure}

We developed two predictive models to evaluate the ability of GPT-4 to map ontology terms to their corresponding ontology IDs:

\begin{enumerate}
    \item \textit{Baseline Model:} This model assumes that GPT-4’s likelihood of correctly mapping an ontology term to its corresponding ID is equal to the baseline accuracy rate for that mapping task (see Table \ref{tab:performance_metrics}). The baseline accuracy represents the overall proportion of terms correctly mapped across all terms, without considering the prevalence of ontology IDs in the biomedical literature.

    \item \textit{Optimal Model:} This model incorporates a threshold for ontology ID counts, identified using receiver-operating characteristic (ROC) curves to maximize sensitivity and specificity (e.g., Fig. \ref{fig:ROC_GO}). For each mapping task, a prevalence threshold (count-based) was determined. Ontology terms with ID counts above this threshold were predicted to be accurately mapped by GPT-4, while terms below the threshold were predicted to be inaccurately mapped.
\end{enumerate}

These models provide a comparative framework for assessing mapping performance: the baseline model evaluates overall accuracy without additional features, while the optimal model utilizes ontology ID prevalence (counts) to enhance prediction accuracy.

\begin{figure}[t!]
    \centering
    \includegraphics[width=0.95\linewidth]{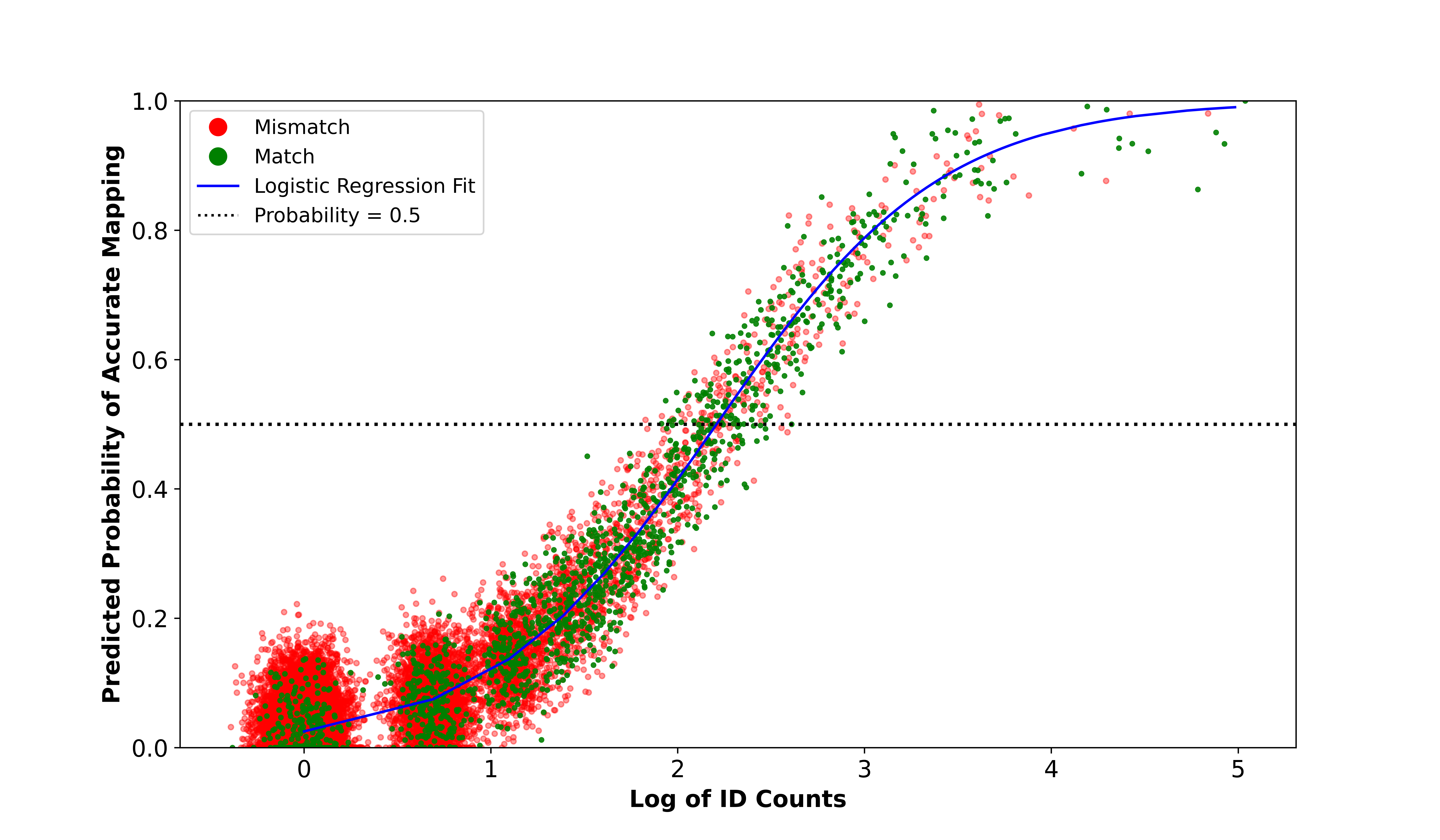}
    \caption{\textbf{Accuracy of Mapping HPO Term to HPO ID Predicted by Logistic Regression.} The log of the ID count in PMC as the predictor (x-axis) was the predictor of accurate mapping (y-axis). Green markers represent correct mappings, and red markers represent incorrect mappings. The blue line shows the logistic regression fit, and the black dashed line represents the threshold probability of 0.5 used to classify mappings as likely accurate or not. Using this threshold, 57\% of the 462 terms that were above the threshold were accurately mapped to their HPO IDs, while only 7\% of 18,338 terms below the threshold were accurately mapped to their HPO IDs. Markers are jittered at lower ID frequencies to enhance visibility on the left side of the plot. Similar models were constructed for the GO and UniProtKB terminologies (not shown).}
    \label{fig:LR_predicted_match}
\end{figure}

For both models, we calculated precision, recall, F1 score, and accuracy using standard methods (see Table \ref{tab:performance_metrics}). Additionally, we developed logistic regression models to predict the probability of accurate mapping of ontology terms to their ontology IDs based on the log-transformed count of their ontology ID counts in the biomedical literature (Fig. \ref{fig:LR_predicted_match}).

\section{Results and Discussion}
\normalsize
We evaluated the accuracy of GPT-4 across four ontology term-to-ontology ID mapping tasks: HPO terms  $\mapsto$ IDs, GO terms $\mapsto$ GO IDs, protein names $\mapsto$ UniProtKB accession numbers, and protein names $\mapsto$ HUGO gene symbols. Mapping accuracy ranged from 8\% for HPO terms to HPO IDs to 95\% for protein names to gene symbols (Table \ref{tab:mapping_performance}). Ontology terms were well represented in the PMC full-text database, with mean counts exceeding 19,000 per term for all terminologies. However, ontology ID frequencies in PMC were much lower, with mean counts of 1, 17, 29, and 11,366 for HPO IDs, GO IDs, UniProtKB accession numbers, and HUGO gene symbols, respectively. Notably, gene symbols had far greater representation in PMC than the other ontology IDs.

% \begin{table}[ht]
% \centering
% \caption{Baseline and ROC-Based Optimal Models Across Different Ontology Mappings}
% \label{tab:mapping_performance}
% \begin{threeparttable}
% \begin{tabular}{lcccc}
% \hline
% \textbf{Model} & \textbf{Precision} & \textbf{Recall} & \textbf{F1} & \textbf{Accuracy} \\
% \hline
% \multicolumn{5}{l}{\textbf{HPO term to ID}} \\
% Baseline & 0.09 & 0.09 & 0.09 & 0.86 \\
% Optimal (CP = 3) & 0.41 & 0.55 & 0.47 & 0.90 \\
% \hline
% \multicolumn{5}{l}{\textbf{GO term to ID}} \\
% Baseline & 0.31 & 0.31 & 0.31 & 0.55 \\
% Optimal (CP = 5) & 0.75 & 0.78 & 0.77 & 0.85 \\
% \hline
% \multicolumn{5}{l}{\textbf{Protein name to AC}} \\
% Baseline & 0.46 & 0.46 & 0.46 & 0.50 \\
% Optimal (CP = 7) & 0.60 & 0.87 & 0.71 & 0.67 \\
% \hline
% \multicolumn{5}{l}{\textbf{Protein name to GS}} \\
% Baseline & 0.95 & 0.95 & 0.95 & 0.90 \\
% Optimal (CP = 32) & 0.95 & 0.99 & 0.97 & 0.95 \\
% \hline
% \end{tabular}
% \begin{tablenotes}
% \footnotesize
% \item The cutpoint (CP) was used by the model to optimize the trade-off between recall and precision.
% \item HPO: Human Phenotype Ontology; GO: Gene Ontology; AC: Accession Code; GS: Gene Symbol.
% \item The baseline model predicts mapping accuracy based on baseline mapping accuracy.
% \item Due to class imbalance, F1 is a better indicator of model performance than accuracy which is biased towards the baseline model due to class imbalances. Note that based on F1, the optimal model outperforms the baseline model on all mappings except protein name-to-gene symbol.
% \end{tablenotes}
% \end{threeparttable}
% \end{table}

\begin{table}[ht]
\centering
\caption{Baseline and Optimal Models Across Ontology Mappings}
\label{tab:mapping_performance}
\begin{threeparttable}
\begin{tabular}{lrrrr}
\hline
\textbf{Model} & \textbf{Prec.} & \textbf{Rec.} & \textbf{F1} & \textbf{Acc.} \\
\hline
\multicolumn{5}{l}{\textbf{HPO term to ID}} \\
Baseline & 0.08 & 1.00 & 0.15 & 0.92 \\
Optimal (ROC=2) & 0.31 & 0.68 & \textbf{0.43} & 0.86 \\
\hline
\multicolumn{5}{l}{\textbf{GO term to ID}} \\
Baseline & 0.33 & 1.00 & 0.49 & 0.67 \\
Optimal (ROC=4) & 0.72 & 0.83 & \textbf{0.77} & 0.84 \\
\hline
\multicolumn{5}{l}{\textbf{Protein name to AC}} \\
Baseline & 0.46 & 1.00 & 0.63 & 0.53 \\
Optimal (ROC=1) & 0.66 & 0.75 & \textbf{0.70} & 0.70 \\
\hline
\multicolumn{5}{l}{\textbf{Protein name to GS}} \\
Baseline & 0.95 & 1.00 & 0.97 & 0.95 \\
Optimal (ROC=746) & 0.96 & 0.81 & 0.98 & 0.78 \\
\hline
\end{tabular}
\begin{tablenotes}
\footnotesize
\item \textbf{Abbreviations:} Prec. = Precision, Rec. = Recall, Acc. = Accuracy, HPO = Human Phenotype Ontology, GO = Gene Ontology, AC = Accession Code, GS = Gene Symbol.
\item The ROC threshold is the ID count in PMC that maximizes sensitivity and specificity. The \textbf{bolded values} indicate the optimal model outperforming the baseline in most cases.
\end{tablenotes}
\end{threeparttable}
\end{table}

Pearson correlation coefficients between ontology term counts in PMC and mapping accuracy were small and accounted for little variance (Table \ref{tab:ROC and Correlation Metrics}). In contrast, correlations between ontology ID counts in PMC and mapping accuracy were higher, accounting for meaningful variance in mapping accuracy for HPO IDs, GO IDs, and UniProtKB accession numbers. No significant correlation was observed between gene symbol counts and mapping accuracy for protein names to gene symbols.

\begin{figure}[b!]
    \centering
\includegraphics[width=0.95\linewidth]{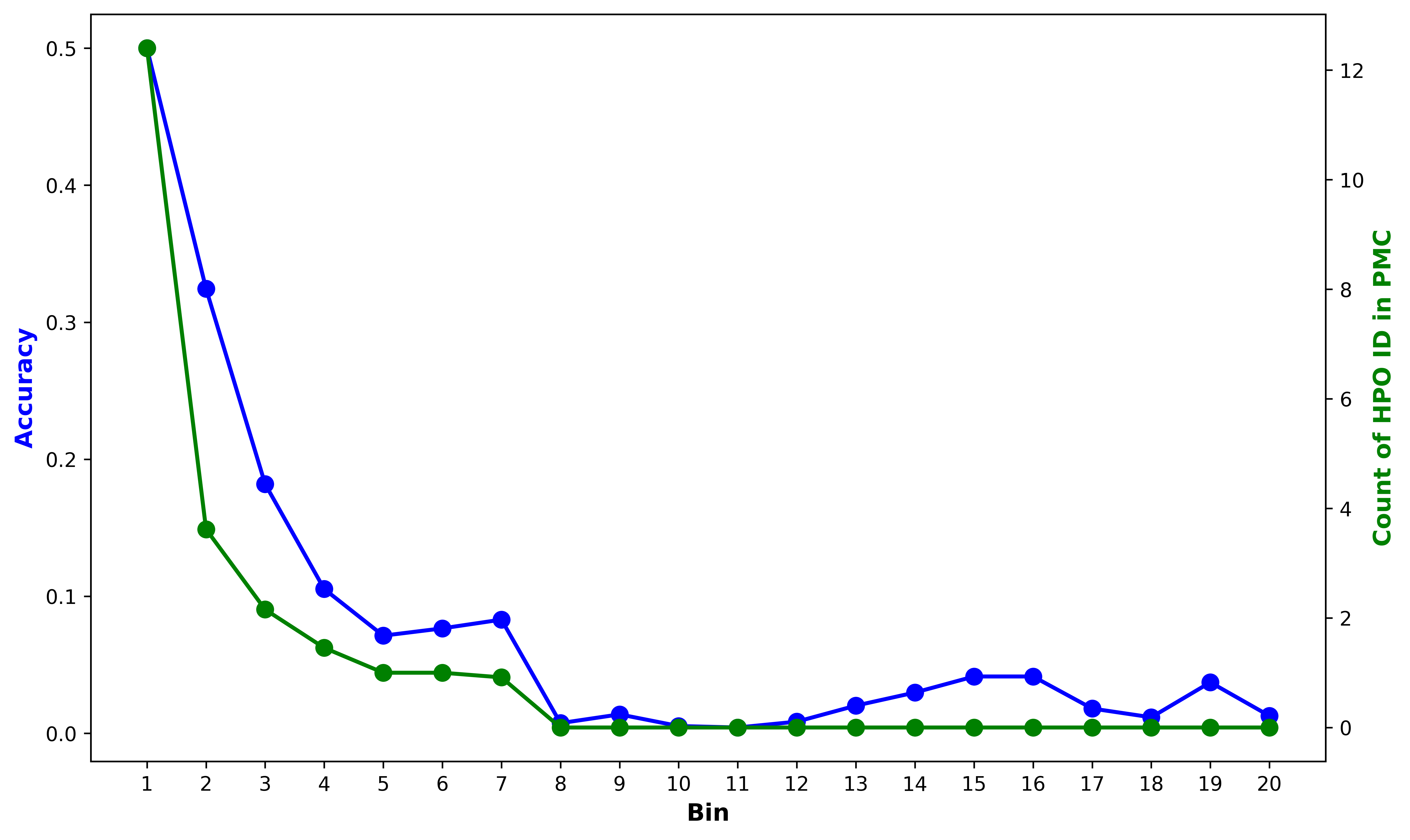}
    \caption{\textbf{Accuracy of Mapping HPO Terms to HPO IDs.} GPT-4 mapped 18,880 terms to their HPO IDs. Terms were rank-ordered by the count of their ontology IDs in the PMC. Terms were divided into 20 equal-sized bins with Bin 1 containing the terms with the highest HPO ID counts in the PMC. For each bin, the mean accuracy (y1-axis) was calculated and the mean HPO ID count was calculated (y2-axis). Note that only terms in the highest ID count bin (Bin 1) were mapped to their HPO IDs accurately.  The remaining 19 bins showed high error rates.}
    \label{fig:HPO Accuracy}
\end{figure}

\begin{figure}[ht!]
    \centering
\includegraphics[width=0.95\linewidth]{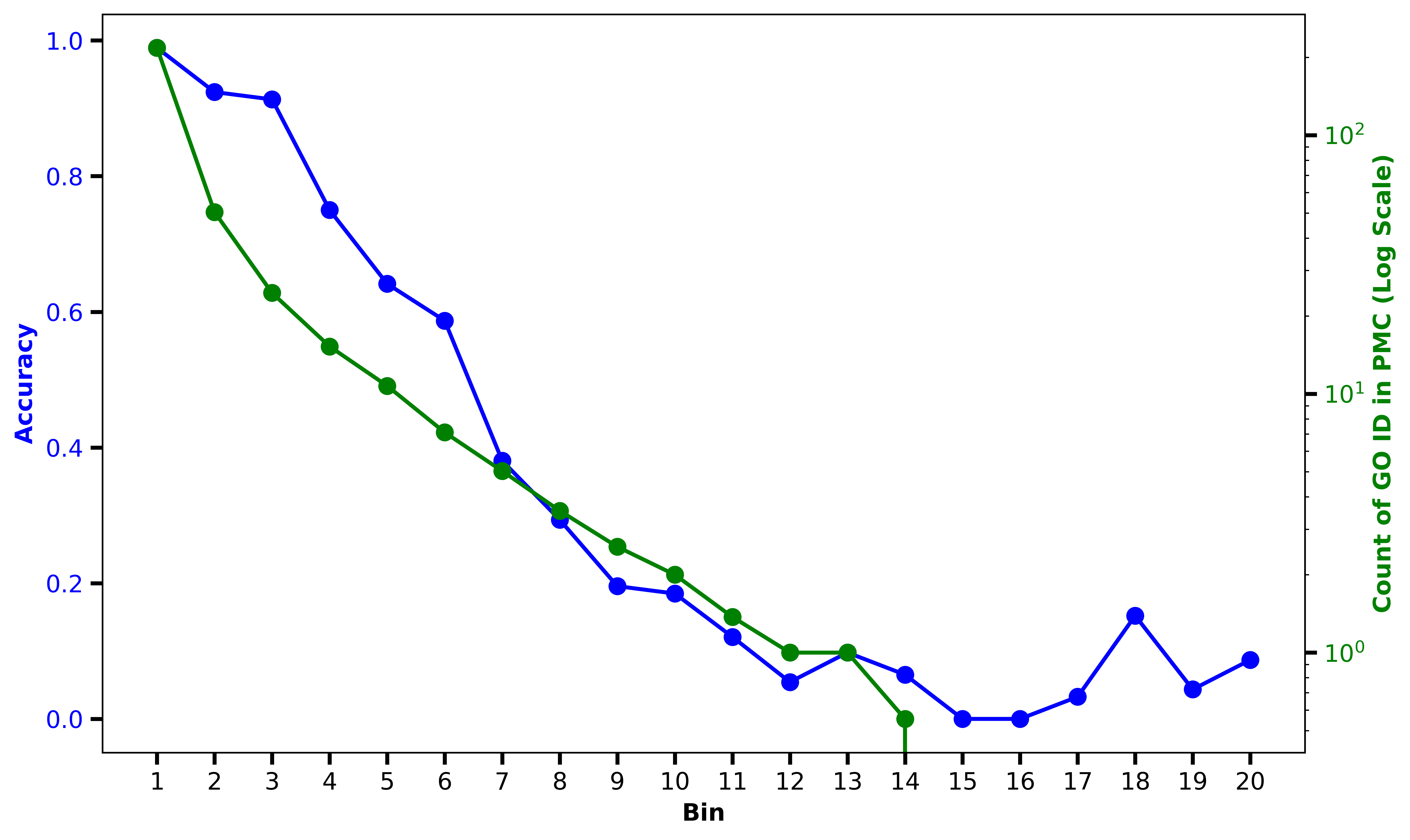}
    \caption{\textbf{Accuracy of Mapping GO Concepts to GO ID.}  GPT-4 mapped 1,839 cellular component terms to their GO IDs with 35\% accuracy. GO terms were ranked according to counts of their GO ID in the PMC. Accuracy and PMC GO ID counts declined steadily and in tandem from bin 1 (highest) to bin 20 (lowest).}
    \label{fig:GO ID bin plot}
\end{figure}

\subsection*{Bin Analysis by Ontology ID Counts}
\begin{figure}
    \centering
    \includegraphics[width=0.95\linewidth]{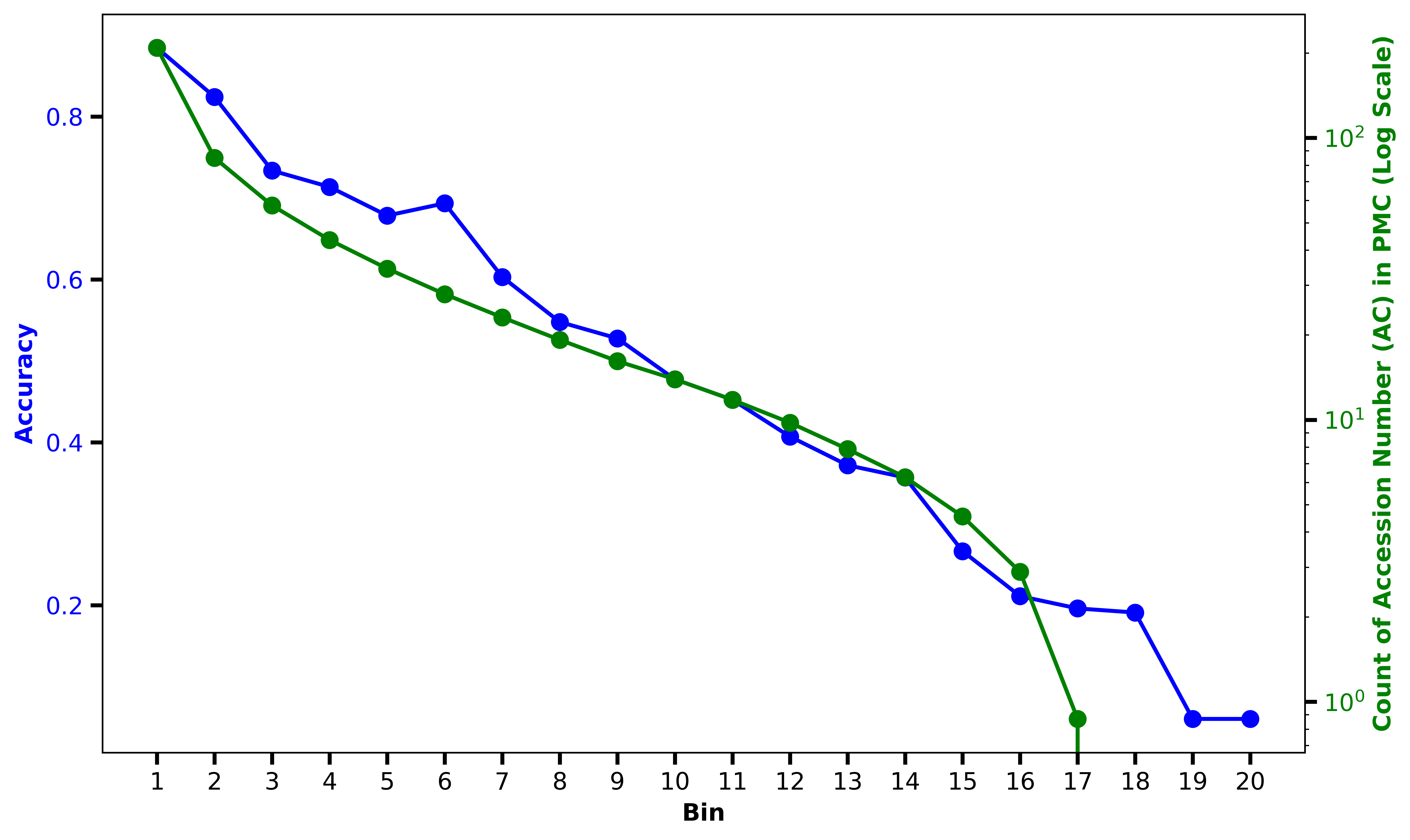}
    \caption{\textbf{Accuracy of Mapping Protein name to Accession Number.} GPT-4 mapped 3,797 protein names to their UniProtKB Accession Numbers with 46\% accuracy.  Protein names were rank-ordered by counts of the Accession Number in the PMC and divided into 20 equal bins. Accuracy and Accession Number frequency (log scale) decline from bin 1 (highest ontology ID counts) to bin 20 (lowest ontology ID counts) in tandem.}
    \label{fig:map_to_AC}
\end{figure}

\begin{figure}[t!]
    \centering
    \includegraphics[width=0.95\linewidth]{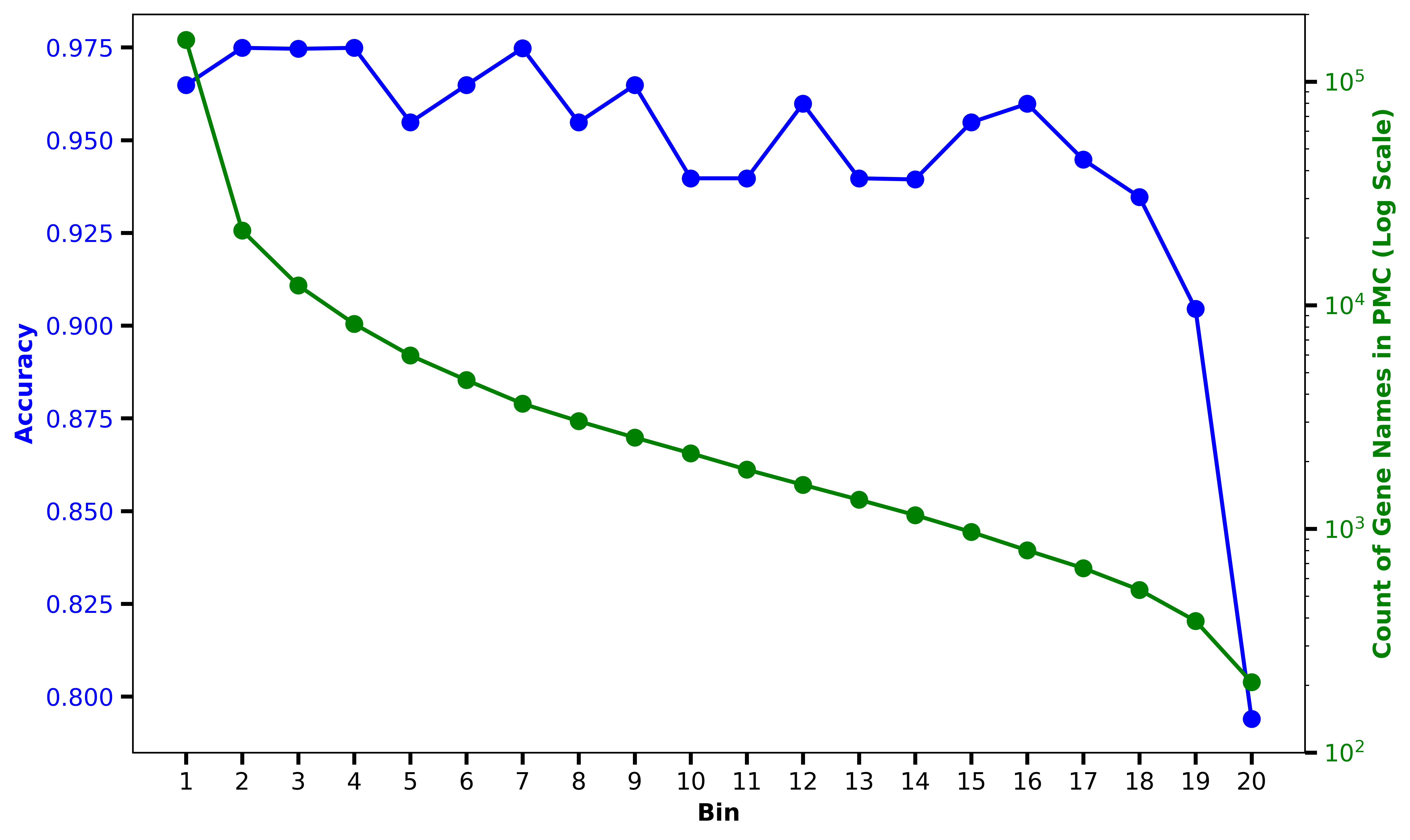}
    \caption{\textbf{Accuracy of Mapping Protein Names to gene symbols by GPT-4.} 3,794 protein names were mapped to their gene symbols by GPT-4. Protein names were rank-ordered by the count of their gene symbols in the PMC full-text database and divided into 20 equal bins based on counts (Bin 1 highest counts).  Accuracy overall was high (95\%) and accuracy (blue-line) and that gene symbol counts in the low low-count bins (18, 19, and 20) are \textgreater 100 and accuracy is \textgreater 80\%.  }
    \label{fig:map_to_GN}
\end{figure}

Ontology terms were rank-ordered and binned according to their corresponding ontology ID counts in PMC (Figs. \ref{fig:HPO Accuracy}, \ref{fig:GO ID bin plot}, \ref{fig:map_to_AC}, and \ref{fig:map_to_GN}). For each bin, we calculated the mean ontology ID cont in PMC and the mean mapping accuracy. A consistent pattern emerged for HPO IDs, GO IDs, and UniProtKB accession numbers: bins with higher ontology ID counts exhibited higher mapping accuracies. However, this pattern did not hold for protein names mapped to gene symbols, where no consistent relationship between ontology ID counts and mapping accuracy was observed (Fig. \ref{fig:map_to_GN}).

\subsection*{Receiver Operating Characteristics (ROC)  Analysis}
To identify optimal frequency thresholds of ID counts in the PMC that were predictive of accurate mappings, ROC curves were created for each mapping task (e.g., Fig. \ref{fig:ROC_GO}). For each ROC curve, we determined the ID count in the PMC that optimized the trade-off between specificity and sensitivity (Table \ref{tab:ROC and Correlation Metrics}). The predictive performance of the count-based threshold model was compared to a baseline model that predicted accuracy based on baseline accuracy rate. Due to class imbalances, the F1 score was used as the primary metric for evaluating model performance. 

The count-based optimal threshold models outperformed the baseline models for HPO IDs, GO IDs, and UniProtKB accession numbers. For these mappings, ontology ID counts in PMC were a meaningful predictor of mapping accuracy. However, this was not the case for mapping protein names to HUGO gene symbols, where the baseline model and count-based models showed no significant differences in performance (Table \ref{tab:performance_metrics}).
\subsection*{Lexicalization of Gene Symbols}
Ontology ID frequency in PMC did not predict mapping accuracy for gene symbols. This anomaly can be explained by several factors:
\begin{itemize}
    \item \textit{High Prevalence of Gene Symbols}. Gene symbols are highly represented in PMC, with a mean frequency of 11,366 per symbol—far greater than the frequencies of HPO IDs, GO IDs, or accession numbers.
    \item \textit{Semantic Design of Gene Symbols}. Unlike arbitrary machine codes (e.g., HPO IDs, GO IDs, or accession numbers), HUGO gene symbols are designed to enhance human recall (e.g., "NEFL" for neurofilament light chain).
    \item \textit{Lexicalization of Gene Symbols}.  Gene symbols are widely used as shorthand for longer gene names (e.g., "GFAP" for glial fibrillary acidic protein). Over time, they have become "lexicalized" \cite{lehmann2002new,stekauer2005handbook, BAKKEN2006106}, acquiring quasi-word status with semantic content beyond their role as identifiers.
\end{itemize}

These factors likely explain why GPT-4 achieved a high baseline accuracy (95\%) for mapping protein names to gene symbols, regardless of gene symbol frequency in PMC (Fig. \ref{fig:map_to_GN}).

\begin{figure}
    \centering
    \includegraphics[width=0.95\linewidth]{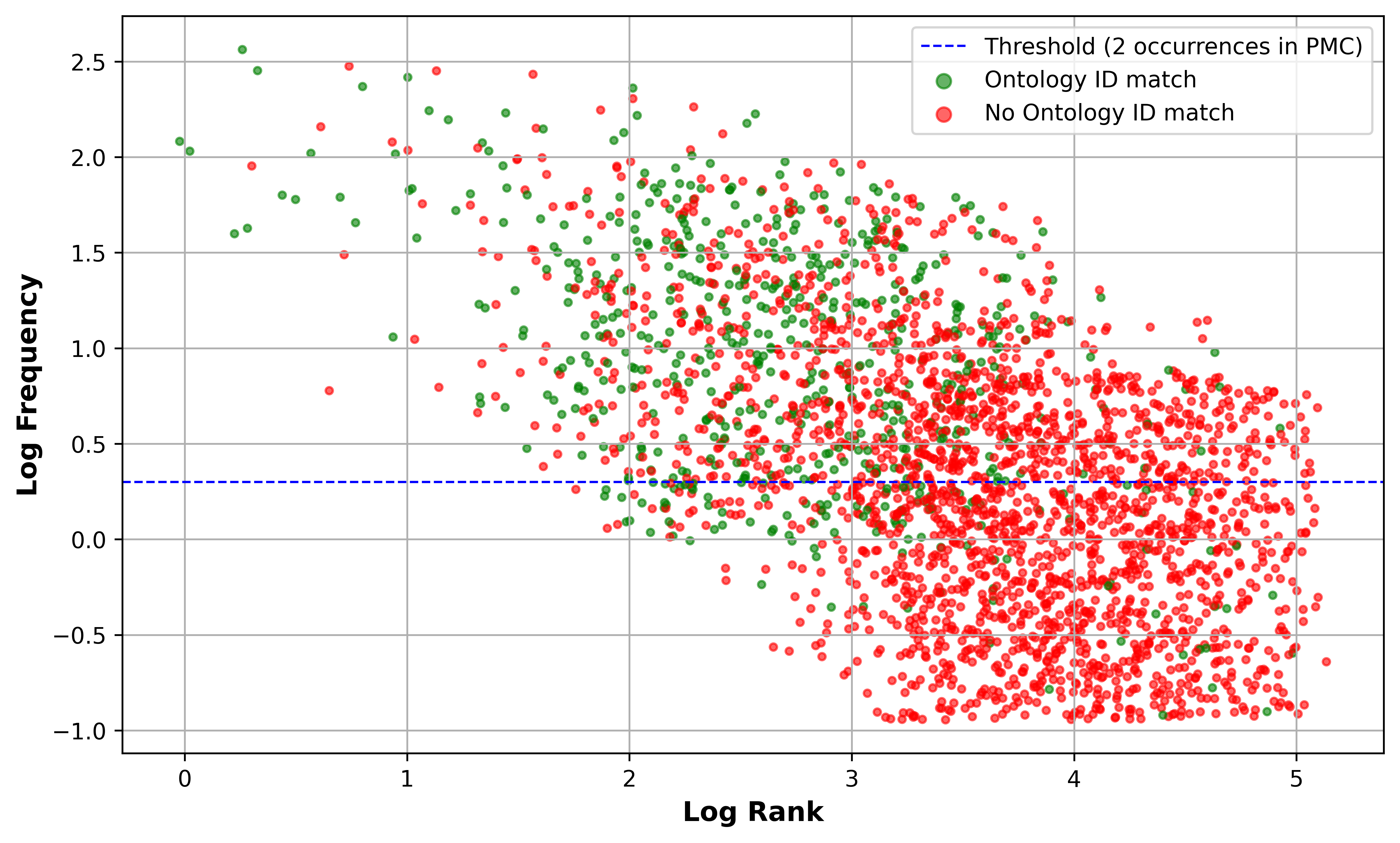}
    \caption{\textbf{Zipf plot of HPO concepts.} GPT-4 mapped 18,800 HPO concepts to their IDs. Green markers (correct mappings) are concentrated in the high-frequency portion (upper left), while red markers (incorrect mappings) dominate the low-frequency range. Both axes are log-transformed. The blue dashed line represents a threshold of two occurrences in PMC, below which mapping accuracy drops significantly. Markers with a rank lower than 1,000 (low frequency terms) were downsampled by 90\% to prevent marker overlap, and jittering was applied to enhance readability in dense regions.}
    \label{fig:Zipf_HPO}
\end{figure}

Given the central role of the HPO in the biomedical literature as one of the most widely used ontologies \cite{robinson2012deep}, we performed additional analyses to examine GPT-4's performance on term-to-ID mapping for this ontology. First, we created a Zipf-style \cite{piantadosi2014zipf, corominas2010universality} log rank-log frequency plot of 18,800 HPO terms, color-coded to indicate whether each term was correctly mapped to its ontology ID by GPT-4 (Fig. \ref{fig:Zipf_HPO}). The plot revealed that a small number of high-frequency, high-rank terms (upper left quadrant) were accurately mapped, whereas the majority of low-frequency, low-rank terms (lower right quadrant) were inaccurately mapped. This distribution highlights the relationship between term frequency and mapping accuracy.
Next, we fitted a logistic regression model to predict the probability of accurate ontology ID mapping based on the frequency of HPO IDs in the PMC database (Fig. \ref{fig:LR_predicted_match}). The resulting logit-probability curve demonstrated a clear trend: terms with low HPO ID frequencies were predominantly predicted to be inaccurately mapped, whereas a small subset of high-frequency terms was predicted to have a high probability of accurate mapping. These findings confirm that HPO ID counts are a predictor of GPT-4's ability to accurately map HPO terms to their ontology IDs. Mapping accuracy sharply increases as ID counts rise.

\subsection*{Implications and Limitations}
The combined results from correlation analysis (Table \ref{tab:ROC and Correlation Metrics}), bin analyses (Figs. \ref{fig:HPO Accuracy}, \ref{fig:GO ID bin plot}, \ref{fig:map_to_AC}, and \ref{fig:map_to_GN}), and ROC curve models indicate that ontology ID counts in PMC are a strong predictor of mapping accuracy for HPO IDs, GO IDs, and UniProtKB accession numbers. However, this relationship did not hold for mapping protein names to gene symbols, likely due to the lexicalization of gene symbols.

This study has several limitations:
\begin{enumerate}
    \item \textbf{Ontology Scope:} While we included all phenotype terms in the HPO, we restricted our analysis of Gene Ontology to the cellular component hierarchy, excluding molecular function and biological process hierarchies. Similarly, for UniProtKB, we focused on the 4,000 most heavily annotated human proteins, leaving the remaining $\sim$16,000 human proteins unexamined.
    \item \textbf{Generalizability:} Extending this analysis to other ontologies, such as RxNorm \cite{RXnorm}, LOINC \cite{mcdonald2003loinc}, and SNOMED CT \cite{lee2014literature}, could yield broader insights. However, large ontologies like SNOMED CT (with over 350,000 concepts) pose significant computational challenges.
    \item \textbf{Training Corpus Assumptions:} The pre-training corpus for GPT-4 is proprietary \cite{naveed2023comprehensive, hadi2024large, chang2024survey}. We used the PMC full-text database \cite{lee2014literature} as a proxy, assuming that IDs absent from PMC are unlikely to appear in GPT-4’s pre-training data.
    \item \textbf{Data Constraints:} Using the PMC API, we obtained raw counts for ontology terms and IDs but could not convert these into frequencies or prevalences due to the absence of dataset size or temporal information.
\end{enumerate}

\section{Conclusions}
Johann Wolfgang von Goethe famously observed, \textit{“Man sieht nur, was man weiß.”} (“We only see what we know.”). This principle aptly describes large language models: their ability to map ontology terms to ontology IDs hinges on whether terms and IDs co-occur during training. Incomplete training data limits performance, suggesting that retrieval-augmented generation \cite{shlyk2024real, garcia2024improving, hier2024simplified} and other supplementary strategies may be necessary to address these gaps.

Our findings also underscore the importance of constructing balanced test datasets when evaluating models that map ontology terms to ontology IDs. Specifically, test sets should represent both high-frequency and low-frequency ontology terms to prevent overestimation of model performance \cite{groza2015automatic, shlyk2024real, garcia2024improving, hier2024simplified, groza2024fasthpocr, groza2024evaluation, lobo2017identifying, weissenbacher2023phenoid}. Underrepresenting low-frequency ontology terms risks obtaining overly-optimistic accuracy metrics that mischaracterize a model’s true capabilities.

\bibliographystyle{IEEEtran}
\bibliography{references}

\end{document}